\documentclass[journal,transmag，review=true]{IEEEtran}

\usepackage[numbers,sort&compress]{natbib}
\usepackage{arydshln}

\usepackage{soul}
\sethlcolor{yellow}
\usepackage{amsmath,amssymb,amsfonts}
\usepackage[doipre={doi:~}]{uri}
\usepackage{pifont}
\usepackage{color}  
\usepackage{graphicx}
\usepackage{subfigure}
\usepackage{threeparttable}
\usepackage{textcomp}
\usepackage{xcolor}
\usepackage{booktabs}
\usepackage{multirow}
\usepackage{algorithm}
\usepackage{algpseudocode}
\usepackage{amsmath}
\usepackage{hyperref}
\usepackage{tabularx}
\usepackage[numbers, sort & compress]{natbib} 
\usepackage{colortbl}
\hypersetup{
colorlinks=true,
linkcolor=blue,
citecolor=blue,
urlcolor=blue 
}
\usepackage{multirow}
\usepackage[normalem]{ulem}
\useunder{\uline}{\ul}{}
\begin{document}

%
\title{PromptGCN: Bridging Subgraph Gaps in Lightweight GCNs}


\author{\IEEEauthorblockN{Shengwei Ji,
Yujie Tian,
Fei Liu*,
Xinlu Li* and Le Wu}

\thanks{This work is partly supported by the National Natural Science Foundation of China under grants 62306100 and 62176085, and the Natural Science Research Project of Anhui Educational Committee under grant 2023AH052180. The authors are equally grateful to the Hefei University Arithmetic Platform for providing support. (\textit{*Corresponding author: Fei Liu and Xinlu Li}).}

\thanks{Shengwei Ji, Yujie Tian and Xinlu Li are with the School of Artificial Intelligence and Big Data, Hefei University,  Hefei, China (e-mail: swji@mail.hfut.edu.cn; tianyujie@stu.hfuu.edu.cn; xinlu.li@hfuu.edu.cn).}

\thanks{Fei Liu and Le Wu are with the School of Computer Science and Information Engineering, Hefei University of Technology, Hefei, China, and also with the Key Laboratory of Knowledge Engineering with Big Data, Ministry of Education, China (e-mail: feiliu@mail.hfut.edu.cn; lewu.ustc@gmail.com).}}

%



\maketitle

\begin{abstract}
Graph Convolutional Networks (GCNs) are widely used in graph-based applications, such as social networks and recommendation systems. 
Nevertheless, large-scale graphs or deep aggregation layers in full-batch GCNs consume significant GPU memory, causing out of memory (OOM) errors on mainstream GPUs (e.g., 29GB memory consumption on the Ogbn-products graph with 5 layers).
The subgraph sampling methods reduce memory consumption to achieve lightweight GCNs by partitioning the graph into multiple subgraphs and sequentially training GCNs on each subgraph.
However, these methods yield gaps among subgraphs, i.e., GCNs can only be trained based on subgraphs instead of global graph information, which reduces the accuracy of GCNs. 
In this paper, we propose PromptGCN, a novel prompt-based lightweight GCN model to bridge the gaps among subgraphs.
First, the learnable prompt embeddings are designed to obtain global information.
Then, the prompts are attached into each subgraph to transfer the global information among subgraphs.
Extensive experimental results on seven large-scale graphs demonstrate that PromptGCN exhibits superior performance compared to baselines.
Notably, PromptGCN improves the accuracy of subgraph sampling methods by up to 5.48\% on the Flickr dataset.
Overall, PromptGCN can be easily combined with any subgraph sampling method to obtain a lightweight GCN model with higher accuracy.

\end{abstract}




%

\section{Introduction}
Graph Convolutional Networks (GCNs) aggregate the features of the neighboring nodes and efficiently transmit information about the nodes among the layers. This approach have achieved significant performance in various graph tasks, such as link prediction \cite{9955364} and node classification \cite{DBLP:journals/tnn/GongZQLZ23}.

\begin{figure}[!t]
  \centering
  \includegraphics[width=1\linewidth]{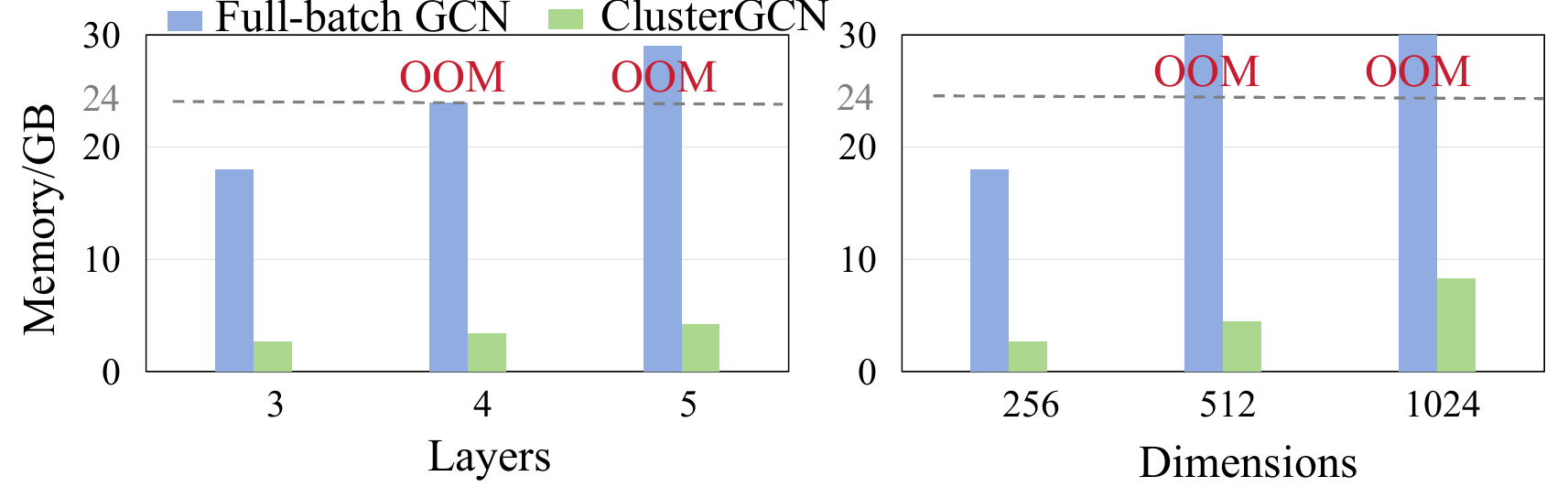}
  \caption{The GPU memory consumption of running full-batch GCN (blue) and ClusterGCN (green) in NVIDIA 3090 GPU on Obgn-products dataset. Full-batch GCN causes an out of memory (OOM) error with the number of layers and the hidden dimension increase. In contrast, the subgraph sampling method has lower memory consumption than full-batch GCN.}
  \label{F111}
\end{figure}

Recent studies indicate that increasing the scale of training batches correlates with higher accuracy in GCNs
\cite{DBLP:conf/mlsys/JiaLGZA20}.
For example, the full-batch approaches \cite{DBLP:conf/iclr/KipfW17}, \cite{DBLP:conf/iclr/VelickovicCCRLB18}, \cite{DBLP:conf/icml/ChenWHDL20}, \cite{DBLP:conf/iclr/Brody0Y22} utilize the global graph information (i.e., the relationships and features among all nodes) to optimize the accuracy of GCNs.
Nevertheless, the memory consumption also rises with the scale of batch size and graph increasing.
For instance, as illustrated in Fig. \ref{F111}, training full-batch GCNs on the large-scale Obgn-products graph--comprising over 20 million nodes and 100 million edges--using an NVIDIA 3090 GPU, results in an out of memory (OOM) error when the number of layers exceeds 3 or the dimensions surpass 512.

To reduce memory consumption, existing studies train GCNs in smaller batches through various sampling methods. Based on different sampling granularities, existing sampling methods can be categorized into three types, detailed as follows.
1) \textit{The node-wise sampling methods} \cite{DBLP:conf/nips/HamiltonYL17}, \cite{DBLP:journals/nn/ZhaoGYH23} 
sample a fixed or random number of neighbors for each node.
Nevertheless, as the number of layers increases, the number of sampled neighboring nodes grows exponentially, leading to the neighbor explosion problem.
2) \textit{The layer-wise sampling methods} \cite{DBLP:conf/iclr/ChenMX18}, \cite{DBLP:conf/nips/Huang0RH18}, \cite{DBLP:conf/nips/LiuW00YS020}, \cite{DBLP:conf/nips/BalinC23} 
sample a fixed number of nodes at each layer to mitigate the neighbor explosion problem.
On the other hand, information loss from unsampled nodes in each layer will result in a reduction in GCN accuracy.
3) \textit{The subgraph sampling methods} \cite{DBLP:conf/kdd/ChiangLSLBH19}, \cite{DBLP:conf/iclr/ZengZSKP20}, \cite{DBLP:conf/iclr/ShiL023}, \cite{DBLP:conf/cikm/XinSHHQ022}, \cite{DBLP:conf/asplos/YangZD023}, \cite{DBLP:conf/eurosys/HuangZ0WJZZZYS24} partitions the global graph into multiple subgraphs, then GCN is trained on each subgraph sequentially.
For instance, ClusterGCN \cite{DBLP:conf/kdd/ChiangLSLBH19} uses a node clustering algorithm METIS \cite{DBLP:journals/siamsc/KarypisK98} to partition the global graph into subgraphs. 
GraphSAINT \cite{DBLP:conf/iclr/ZengZSKP20} designs the sampling probabilities for nodes and edges and uses these to sample the subgraph in each batch.
Compared with the node- and layer-wise sampling methods, the subgraph sampling methods reduces the memory consumption of GCN while ensuring that all nodes are sampled.
Fig. \ref{F111} illustrates that the subgraph sampling method ClusterGCN \cite{DBLP:conf/kdd/ChiangLSLBH19} has lower memory consumption than full-batch GCN.

\begin{figure}[!t]
\centering  
\subfigure[PromptGCN expands the local receptive field of subgraphs]{
\includegraphics[width=1\linewidth]{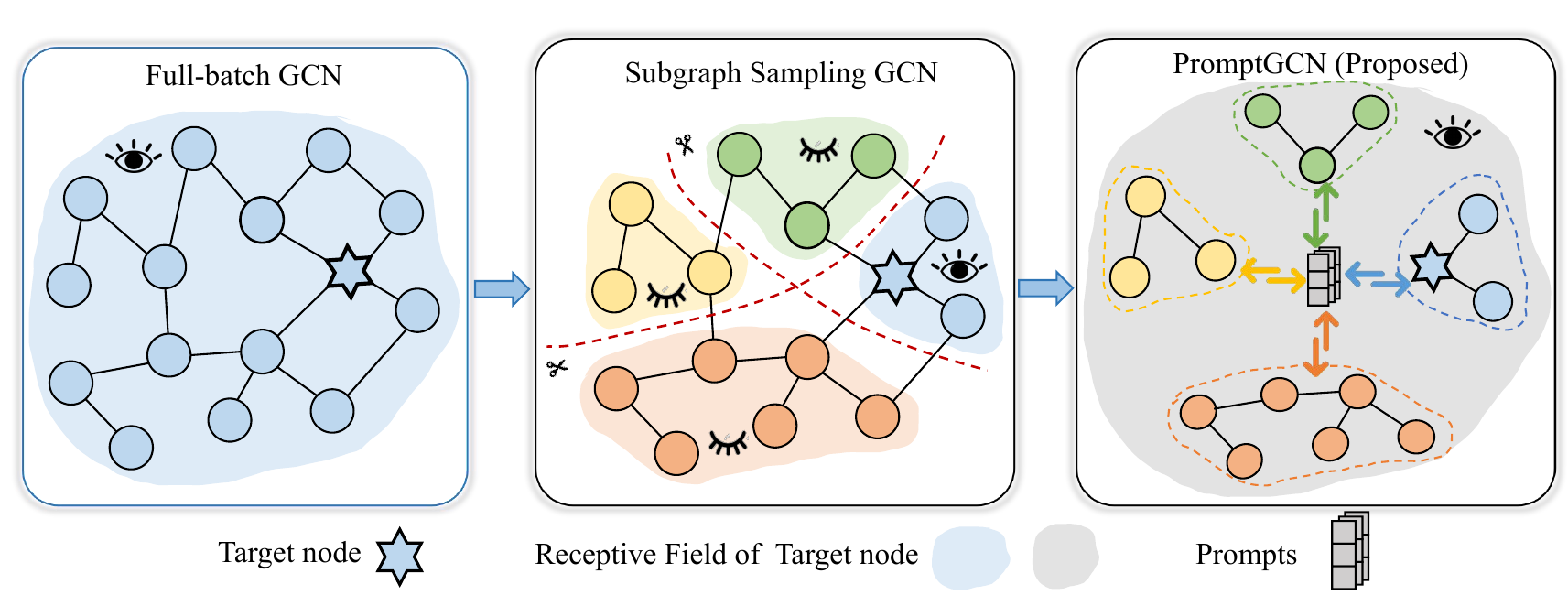}}
\subfigure[Gaps among subgraphs reduce test accuracy]{
\includegraphics[width=0.8\linewidth]{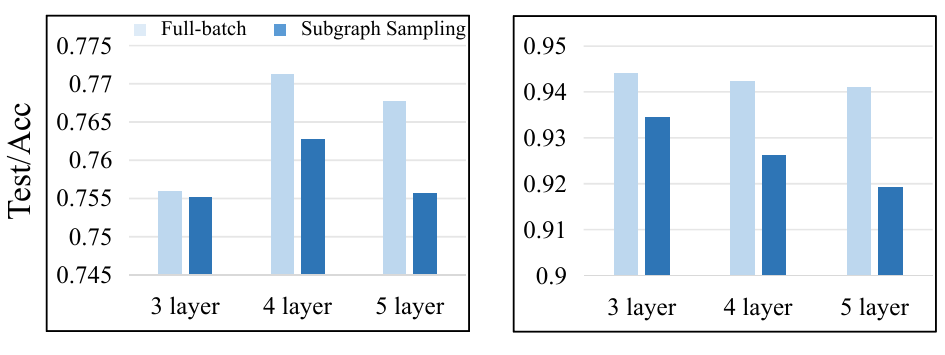}}
\caption{Impact of the local receptive field of subgraphs on test accuracy (Ogbn-products and Reddit datasets).}
\label{F1}
\end{figure}

However, subgraph sampling methods discard information outside the subgraph during the training process of GCN, resulting in only local receptive fields within the subgraph (i.e., the aggregation range of the target node reduced) \cite{kazi2019inceptiongcn}, \cite{DBLP:journals/tkdd/TangLGZWZL23}. This limitation makes it challenging to capture global graph information, as shown in Fig. \ref{F1}(a).
Moreover, subgraph sampling methods yield gaps among subgraphs, which can lead to reduced connectivity and, consequently, decrease the accuracy of GCN.
Fig. \ref{F1}(b) illustrates that the accuracy of the subgraph sampling method is lower than that of the full-batch method across different GCN layers.
Concerning this, few studies optimize subgraph sampling by integrating external information to reduce the gaps among subgraphs. 
For instance, GAS \cite{DBLP:conf/icml/FeyLWL21} and LMC \cite{DBLP:conf/iclr/ShiL023} include the node outside the subgraph into the training process of GCN to expand the subgraphs' receptive fields.
For these methods, as the graph scale and GCN complexity increase, the time and space consumption also increases correspondingly.
Therefore, how to bridge the gap among the subgraphs to improve accuracy while maintaining low memory consumption is an essential problem for GCN.

Recent studies \cite{ding2023parameter}, \cite{DBLP:conf/kdd/SunZHWW22}, \cite{liu2023graphprompt}, \cite{Fang2023UniversalPT} have indicated that prompts are effective in bridging the gap between pre-trained models and downstream tasks.
In this paper, the ``prompt'' strategy is introduced and applied for the first time in subgraph sampling methods for large-scale graphs, with these prompts typically consisting of a small number of free parameters and occupying minimal memory.
Specifically, prompts are used to transfer global information among subgraphs, bridging the gap among each subgraph, which is shown in Fig. \ref{F1}(a).
Nevertheless, there are two challenges in designing prompts for subgraph sampling: 
(1) \textit{How to design prompt embeddings that effectively capture global information?} (2) \textit{How to transfer global information to bridge the gaps among subgraphs?}  
To address the two challenges, we propose a prompt-based lightweight GCN (PromptGCN) model.
First, PromptGCN incorporates learnable prompt embeddings that capture global information beyond the target subgraph. Second, these prompt embeddings are attached with the target subgraph in various ways, effectively transferring global information to the subgraph.
Extensive experimental results indicate that PromptGCN effectively bridges the gaps among subgraphs and enhances the accuracy of GCNs, while maintaining low memory consumption on large-scale graphs.
The main contributions of PromptGCN are as follows:
\begin{itemize}
    \item 
    PromptGCN extends the receptive field of subgraph sampling GCN and bridges gaps between subgraphs by utilizing prompts to capture and share global information.
 

    \item 
    PromptGCN can be easily integrated with any subgraph sampling method to obtain a lightweight GCN model that achieves higher accuracy on multiple downstream tasks.
    
    \item Extensive experiments on seven large-scale datasets across multiple downstream tasks demonstrate that PromptGCN outperforms baselines in terms of accuracy and memory consumption. 
    \end{itemize}

The paper is structured as follows: In Section \ref{preli}, we present the problem formulation for PromptGCN.
Section \ref{model} provides a detailed explanation of the proposed PromptGCN model.
Section \ref{expe} describes the experiments conducted to evaluate the performance of PromptGCN.
Section \ref{related} reviews existing GCN sampling and prompt learning methods.
Finally, Section \ref{con} concludes the paper.

\begin{figure*}[!t]
  \centering
  \includegraphics[width=0.85\linewidth]{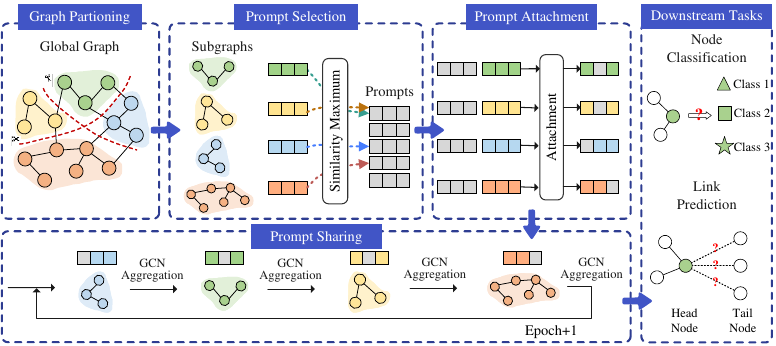}
  \caption{The overall process of PromptGCN includes: Graph Partitioning, Prompt Selection, Prompt Attachment, Prompt Sharing, and PromptGCN for Downstream Taks. First, in the Graph Partitioning module, the global graph is partitioned into several subgraphs. Second, in the Prompt Selection module, the node features within each subgraph adaptively select global prompt embeddings. Third, in the Prompt Attachment module, the selected global prompt embeddings are attached to the node features. Fourth, in the Prompt Sharing module, the processed node-prompt embedding pairs are input into a GCN model that is trained sequentially across subgraphs, sharing global graph information across subgraphs. Finally, PromptGCN is applied to various downstream tasks (e.g., Node Classification and Link Prediction) to guide the model training process.}
  \label{F12}
\end{figure*}

\section{Problem Formulation}\label{preli}
The global graph is defined as $\mathcal{G} = (\mathcal{V}, \mathcal{E}, \mathcal{A})$, where $\mathcal{V} = [{\textbf{\textit{v}}_1, \textbf{\textit{v}}_2, \dots, \textbf{\textit{v}}_n}$] represents the nodes in the global graph, and $\mathcal{E}$ represents the edges. An edge from node $\textbf{\textit{v}}_i$ to node $\textbf{\textit{v}}_j$ is denoted as $(\textbf{\textit{v}}_i, \textbf{\textit{v}}_j) \in \mathcal{E}$, and $\textbf{\textit{v}}_j$: $j \in \mathcal{N}_i$ denotes a neighbor of node $\textbf{\textit{v}}_i$. The adjacency matrix $\mathcal{A}$ indicates whether edges exist between nodes, using 1 to signify their presence and 0 to signify their absence. The input features of the nodes are denoted as $\mathcal{H} = (\textbf{\textit{h}}_1, \textbf{\textit{h}}_2, \dots, \textbf{\textit{h}}_n) \in \mathcal{R}^{n \times d}$, where $n$ is the number of nodes and $d$ is the input feature dimension. For instance, the features of nodes $\textbf{\textit{v}}_i$ and $\textbf{\textit{v}}_j$ are denoted as $\textbf{\textit{h}}_i$ and $\textbf{\textit{h}}_j$, respectively.

Training GCN models on a global graph significantly increases memory consumption. To alleviate this problem, we introduce a graph vertex partitioning method that partitions the input graph data into local subgraphs by cutting edges.
Specifically, the graph partition method partitions the global graph $\mathcal{G}$ into $c$ subgraphs (i.e., $\mathcal{G}_1$,\dots, $\mathcal{G}_c$), and each subgraph is denoted as $\mathcal{G}_t$, where $t = (1,\dots, c)$. 
The nodes in the global graph are correspondingly clustered into $c$ clusters $\mathcal{V}$ = [$\mathcal{V}_1,\dots, \mathcal{V}_c$]. 
We then have $\bigcup_{t=1}^c \mathcal{V}\left(\mathcal{G}_t\right)=\mathcal{V}$ and $\bigcap_{t=1}^c \mathcal{V}\left(\mathcal{G}_t\right)=\emptyset$.
In addition, we can partition the feature matrix $\mathcal{H}$ and the labels $\mathcal{Y}$ into [$\mathcal{H}_1,\dots, \mathcal{H}_c$] and [$\mathcal{Y}_1,\dots, \mathcal{Y}_c$], respectively. $\mathcal{H}_t = (\textbf{\textit{h}}_{1t}, 
\textbf{\textit{h}}_{2t}, \dots, \textbf{\textit{h}}_{nt}) \in \mathcal{R}^{n \times d}$ consists of the node feature matrix in the $t$-th subgraph. 
To enhance the connection between subgraphs, we propose to utilize prompt embedding $\mathcal{P}$ = [$\mathcal{P}_1,\dots, \mathcal{P}_M$] to obtain global graph information, where $M$ is the number of prompt embeddings.
The node features of the $t$-th subgraph $\mathcal{H}_t$, and the shared prompt matrix $\mathcal{P}$ are fed into the GCN model, which is trained sequentially from the 1st subgraph to the $c$-th subgraph. The shared prompt embedding $\mathcal{P}$ is passed from one subgraph to the next to transfer the global graph information among subgraphs.

\textbf{Problem Definition}. 
Given a global graph $\mathcal{G}$, a node feature matrix $\mathcal{X}$, and a shared prompt embedding $\mathcal{P}$, a graph partitioning method is employed to partition $\mathcal{G}$ into multiple subgraphs. Our goal is to train the GCN based on subgraph to reduce memory consumption while sharing global graph information through the shared prompt embedding $\mathcal{P}$ to enhance GCN accuracy on different downstream tasks. Ultimately, we aim to achieve high GCN accuracy with minimal memory consumption.

\section{Model Description}\label{model}
Subgraph sampling methods discard information outside the subgraph during GCN training, creating gaps among subgraphs that can reduce connectivity. Therefore, we propose a prompt-based lightweight GCN (PromptGCN) to learn global information and bridge the gaps among subgraphs.

The framework of PromptGCN is outlined in Section \ref{s4.1}. Section \ref{s4.2} covers Graph Partition, while Section \ref{s4.3} details Prompt Selection. Prompt Attachment is discussed in Section \ref{s4.4}, and Prompt Sharing in Section \ref{s4.5}. Finally, the application of PromptGCN for Downstream Tasks is presented in Section \ref{s4.6}.

\subsection{PromptGCN Framework}\label{s4.1}

The overall framework of PromptGCN is illustrated in Fig. \ref{F12}, comprising Graph Partitioning, Prompt Selection, Prompt Attachment, Prompt Sharing, and Prompts for Downstream Tasks. 
First, the graph partitioning algorithm partitions the global graph $\mathcal{G}$ into several subgraphs, each containing the feature matrices of the nodes $\mathcal{H}_t$. The input to PromptGCN consists of the node feature matrix $\mathcal{H}_t$ and a small set of shared prompt embedding matrices $\mathcal{P}$ (detailed in Section \ref{s4.2}). 
Second, each node feature adaptively selects the most relevant prompt embedding and attaches the selected prompt embedding to itself (detailed in Section \ref{s4.3} and \ref{s4.4}).
Third, the processed node-prompt embedding pairs are then fed into the GCN model, which is trained sequentially across subgraphs, with the shared prompt embeddings being passed from one subgraph to the next (detailed in Section \ref{s4.5}). 
Finally, PromptGCN is applied to various downstream tasks (e.g., Node Classification and Link Prediction) to guide the model training process (detailed in Section \ref{s4.6}).
In summary, the prompt embeddings shared among subgraphs carry crucial global information, which is effectively transmitted to different subgraphs through the interaction between node features and prompt embeddings. The details are in Algorithm \ref{Al1}.

\begin{algorithm}[!t]
  \caption{The PromptGCN Process} \label{Al1}\small
  \begin{algorithmic}[1]
    \Require
      Global graph $\mathcal{G} = (\mathcal{V}, \mathcal{R})$,
      initialization of prompt embeddings $\mathcal{P}$,
      the feature matrix of nodes $\mathcal{V}$ is denoted as $\mathcal{H}$.
      \State \textbf{Initialize} the node feature matrix $\mathcal{H}$.
       \State \textbf{Initialize} the shared prompt embedding matrix $\mathcal{P}$.
      \State Partition the global graph $\mathcal{G}$ into $c$ local subgraphs, where the node feature matrix $\mathcal{H}$ is partitioned into $\mathcal{H} = [\mathcal{H}_1,\dots, \mathcal{H}_c]$.
      $\mathcal{H}_t$ denotes the node features of the $t$-th subgraph, where $t = 1, \dots, c$.
       \For{$t=1,2, \ldots, c$}
      \State The node feature matrix of the $t$-th subgraph is denoted as \State $\mathcal{H}_t$ =  $(\textbf{\textit{h}}_{1t},\dots, \textbf{\textit{h}}_{nt})$.
      \State Calculate the similarity between the target node feature $\textbf{\textit{h}}_{jt}$ \State and the prompt embedding $\mathcal{P}$ in the $t$-th subgraph.
      \State Select the prompt $\mathcal{P}$ to attach to the target node feature \State $\textbf{\textit{h}}_{jt}$.
      \State Feed $\mathcal{H}_t$ and $\mathcal{P}$ into the GCN model.
      \State Compute the subgraph loss on different downstream tasks.
      \State Update node features $\mathcal{H}_t$ and prompt embeddings $\mathcal{P}$. 
      \EndFor
      \State \textbf{Output}: Global graph node $\mathcal{H}$ and shared prompt $\mathcal{P}$
  \end{algorithmic}
\end{algorithm}

\subsection {Graph Partitioning}\label{s4.2}
Training a GCN model on a global graph leads to high memory consumption. To address this issue, we employ the graph vertex partitioning approach, which reduces memory consumption by partitioning the global graph into multiple local subgraphs.

Specifically, the graph partitioning method partitions the global graph $\mathcal{G}$ into $c$ local subgraphs, where the nodes
in the graph are allocated into $c$ clusters $\mathcal{V}$ = [$\mathcal{V}_1$, $\mathcal{V}_2$, ..., $\mathcal{V}_c$]. $\mathcal{V}_t$ consists of the nodes in the $t$-th subgraph, where $t = 1, 2,\dots, c$. The details are as follows:
\begin{equation}\label{eq3}
\mathcal{G}=\left[\mathcal{G}_1, \cdots, \mathcal{G}_c\right]=\left[\left\{\mathcal{V}_1, \mathcal{E}_1\right\}, \cdots,\left\{\mathcal{V}_c, \mathcal{E}_c\right\}\right],
\end{equation}
where $\mathcal{V}$ denotes the nodes, and $\mathcal{E}$ represents the edges, with each $\mathcal{E}_t$ consisting solely of edges between nodes in $\mathcal{V}_t$. We then have $\bigcup_{t=1}^c \mathcal{V}\left(\mathcal{G}_t\right)=\mathcal{V}$ and $\bigcap_{t=1}^c \mathcal{V}\left(\mathcal{G}_t\right)=\emptyset$.

Furthermore, we can partition the feature matrix $\mathcal{H}$ and the labels $\mathcal{Y}$ into [$\mathcal{H}_1,\dots, \mathcal{H}_c$] and [$\mathcal{Y}_1,\dots, \mathcal{Y}_c$], respectively, based on the partition [$\mathcal{V}_1,\dots, \mathcal{V}_c$], where $\mathcal{H}_t$ and $\mathcal{Y}_t$ consist of the features and labels of the nodes in $\mathcal{V}_t$.

\subsection{Prompt Selection}\label{s4.3}
Subgraphs discard information outside their boundaries during training process, making it difficult for nodes within the subgraph to obtain global information representations. To address this, we adopt prompt embedding $\mathcal{P}$ to learn global information and enable node features to autonomously select the appropriate global information.

Specifically, let the prompt embedding $\mathcal{P}$ consist of a series of learnable embeddings, $\mathcal{P} = [\mathcal{P}_1, \dots, \mathcal{P}_M]$, where $M$ represents the number of prompt embeddings. 
The input features of the nodes in the $t$-th subgraph are denoted as $\mathcal{H}_t = (\textbf{\textit{h}}_{1t}, \textbf{\textit{h}}_{2t}, \dots, \textbf{\textit{h}}_{nt}) \in \mathcal{R}^{n \times d}$, where $t = 1, 2, \dots, c$, $n$ is the number of nodes, and $d$ is the input feature dimension. The embedding dimension of each prompt embedding $\mathcal{P}_m \in \mathcal{R}^d$ is aligned with the node feature dimension $\textbf{\textit{h}}_{jt} \in \mathcal{R}^d$, where $m = 1, 2, \dots, M$ and $j = 1, 2, \dots, n$. The similarity between node feature $\textit{\textbf{h}}_j$ and prompt embedding $\mathcal{P}_m$ is computed by the dot product, and the $\mathcal{P}_{m}$ with the largest association with the node feature is selected as the prompt message for that node:
\begin{equation}\label{eq3}
\mathcal{P}_{{m} }=\arg\max(\sum_{{m}=1}^{M}\mathcal{P}_{m} \cdot \textit{\textbf{h}}_{jt}^T),
\end{equation}
where $\mathcal{P}_{m}$ is the adaptively chosen prompt embedding for the node feature. After selecting the appropriate prompt embedding, determining how to attach it to the subgraph becomes another key issue to address.

\subsection{Prompt Attachment}\label{s4.4} 
There are two main types of feature fusion methods in deep learning, i.e., concatenate and point-wise operations. Therefore, we use two types of approaches to fuse node features and prompt embeddings, allowing nodes to acquire global information.
The specific methods are listed below: 
(1) The first method directly concatenates node features with prompt embeddings; 
(2) The second method point-wise additions the node features to the prompt embeddings; 
(3) The third method point-wise multiplies the node features with the prompt embeddings; 
(4) The fourth method point-wise additions prompt embeddings to node features with specific weights. 
In the above scheme, (1) is a concatenate operation, while (2), (3), and (4) are point-wise operations.
The corresponding formulas are provided below:
\begin{equation}\label{eq4}
\textit{\textbf{h}}_{jt}=\left\{\begin{array}{cl}
Concat (\textit{\textbf{h}}_{jt}, \mathcal{P}_{m}) \\
\textit{\textbf{h}}_{jt}+ \mathcal{P}_{m}\\
\textit{\textbf{h}}_{jt}* \mathcal{P}_{m}\\
\alpha \textit{\textbf{h}}_{jt}+ (1-\alpha )\mathcal{P}_{m}
\end{array}\right.,
\end{equation}
where $\alpha$ is a custom parameter that controls the weight of prompts added to the node feature. Through Experiment \ref{5.6}, we found that using the concatenate operation to attach prompt embeddings to node features maximize the delivery of global information. Therefore, in this paper, the concatenate operation is used as the main attachment method.

\subsection{Prompt Sharing}\label{s4.5} 
Based on the local subgraph, GCN is able to capture the structured feature of nodes and their neighbors in the graph. At each layer, GCN employs a message passing and aggregation mechanism, where it collects features from neighboring nodes and efficiently propagates node information across layers using various linear combination functions. The aggregation process for passing node features $\textbf{\textit{h}}_{jt}$ in GCN at the $k$-th layer is represented as:
\begin{equation}\label{eq1}
\textit{\textbf{h}}_{it}^k=f\left(\textbf{\textit{h}}_{it}^{k-1},\left\{\textbf{\textit{h}}_{jt}^{k-1}: {jt} \in \mathcal{N}_{it}\right\} ; \theta^k\right), 
\end{equation}
where $\mathcal{N}_{it}$ is the neighbors of node feature $\textbf{\textit{h}}_{it}$ and $\theta^k$ is the $k$-th layer learnable GCN parameter, including node feature matrix $\mathcal{H}_t$ and shared prompt embedding $\mathcal{P}$. $f$ denotes the aggregation function, which is used to pass the neighbor node features to the target node feature $\textbf{\textit{h}}_{it}\in\mathcal{R}^{d}$, where $i = 1, 2,\dots, n$ and $t = 1, 2,\dots, c$.

The GCN is trained sequentially across subgraphs, with the shared prompt embeddings $\mathcal{P}$ passed from one subgraph to the next, thereby sharing global information across subgraphs.

\subsection{PromptGCN for Downstream Tasks}\label{s4.6}
Combining the GCN model with downstream tasks allows for effective training of the prompt parameters and refinement of the global information carried within the prompts. Therefore, we apply the GCN model across different downstream tasks.

\textbf{Node classification.} 
The node classification task consists of $\mathcal{|Y|}_y)$ classes, and $y_{m} \in \mathcal{Y}_t$ is the ground label of the target node. Specifically, we predict the classification labels for target node feature $\textit{\textbf{h}}_{it}$. The optimized classification loss is described as:
\begin{equation}\label{eq6}
\mathcal{L}_t=\frac{1}{\left|\mathcal{Y}_t\right|} \sum_{m \in \mathcal{Y}_t} f\left(y_{m}, \textit{\textbf{h}}_{it}\right)+\gamma\mathcal{L}_o, 
\end{equation}
where $\gamma$ is a customized hyper-parameter, and $f$ is the classification loss function. 
Through Experiment \ref{5.6}, we found a correlation between the number of prompts and the classes in the node classification task. Specifically, model performance is optimized when the number of prompts is similar to the number of class. Since the classes are independent of each other, we introduced a regularization term $\mathcal{L}_o$ to ensure a larger difference between different $\mathcal{P}$.
\begin{equation}\label{eq5}
\mathcal{L}_o=\sum_{|\mathcal{Y}_t|}\left\|\mathcal{P}*\left(\mathcal{P}\right)^{\top}-I\right\|_F^2,
\end{equation}
where $I$ is the identity matrix. Prompt embeddings $\mathcal{P}$ corresponding to different classification labels should be distinct from one another, meaning that these prompt embeddings should exhibit low similarity to each other.

\textbf{Link Prediction.} The GCN model predicts the score between the head node feature $\textbf{\textit{h}}_{it}\in\mathcal{R}^{d}$ and the tail node feature $\textbf{\textit{h}}_{jt}\in\mathcal{R}^{d}$ by defining a scoring function $ f\left(\textit{\textbf{h}}_{it}, \textit{\textbf{h}}_{jt}\right)$. In addition, GCN improves the prediction accuracy by randomly negatively sampling some incorrect samples of tail node feature $\textbf{\textit{h'}}_{jt}\in\mathcal{R}^{d}$. The loss function equation is as follows:
\begin{equation}
\mathcal{L}_t=-\frac{1}{N} \sum_i\left(\log \left(f\left(\textit{\textbf{h}}_{it}, \textit{\textbf{h}}_{jt}\right)\right)+ \log \left(1-(f\left(\textit{\textbf{h}}_{it}, \textit{\textbf{h}'}_{jt}\right))\right)\right)\label{eq22},
\end{equation}
where $N$ represents the number of links.

\section{Experiments}
\label{expe}
In this section, we introduce the experimental setup (Section \ref{5.1}) and proceed to answer the following research questions and validate the model's contribution in Sections \ref{5.2}–\ref{5.6}:

\begin{itemize}
\item RQ1: How does the performance of PromptGCN on node classification and link prediction tasks compare to the baselines?

\item RQ2: How does the memory consumption of PromptGCN compare to that of a full-batch model?




\item RQ3: How do hyperparameters affect the performance of PromptGCN?
\end{itemize}

\begin{table}[!t]
\centering
\caption{Statistics of the node classification (above) and link prediction (below) datasets used in this paper.}
\scalebox{0.88}{
\begin{tabular}{ccccc}
\hline
Dataset                                                        & $\#$Nodes            & $\#$Edges            & $\#$Lables           & Metrics           \\ \hline
\begin{tabular}[c]{@{}c@{}}Amazon\\ CoBuyPhoto\end{tabular}    & 7,650                & 238,163              & 8                    & ACC \& F1             \\
\begin{tabular}[c]{@{}c@{}}Amazon\\ CoBuyComputer\end{tabular} & 13,752               & 491,722              & 10                   & ACC \& F1             \\
Flickr                                                         & 89,250               & 899,756              & 7                    & ACC \& F1             \\
Reddit                                                         & 232,965              & 114,848,857          & 41                   & ACC \& F1             \\
Ogbn-products                                                  & 2,449,029            & 126,067,309          & 47                   & ACC \& F1             \\ \hline
\multicolumn{1}{l}{}                                           & \multicolumn{1}{l}{} & \multicolumn{1}{l}{} & \multicolumn{1}{l}{} & \multicolumn{1}{l}{} \\ \hline
Dataset                                                        & $\#$Nodes            & $\#$Edges            & $\#$Node Deg.        & Metric           \\ \hline
Ogbl-citation2                                                 & 2,927,963            & 30,561,187           & 20.7                 & MRR                  \\
Ogbl-collab                                                    & 235,868              & 1,285,465            & 8.2                  & Hits@100              \\ \hline
\end{tabular}}\label{2222}
\end{table}

\subsection{Setup}\label{5.1}
\textbf{Datasets.} To verify the generalization ability of the model across different datasets, we evaluate the PromptGCN model on five large-scale public datasets about node classification tasks: AmazonCoBuyPhoto\footnote{\url{https://github.com/shchur/gnn-benchmark\#datasets}\label{fn:1}}, AmazonCoBuyComputer\footref{fn:1}, Flickr\footnote{\url{https://github.com/GraphSAINT/GraphSAINT}}, Reddit\footnote{\url{http://snap.stanford.edu/graphsage/}}, and Ogbn-products\footnote{\url{https://ogb.stanford.edu/docs/nodeprop/\#ogbn-products}}. Additionally, we evaluate the PromptGCN model on two large-scale public datasets about link prediction tasks: Ogbl-citation2\footnote{\url{https://ogb.stanford.edu/docs/linkprop/\#ogbl-citation2}} and Obgl-collab\footnote{\url{https://ogb.stanford.edu/docs/linkprop/\#ogbl-collab}}. The datasets detail are shown in Table \ref{2222}.

\textbf{Baselines.} In this paper, we provide a brief description of the full-batch and subgraph sampling models.
\begin{itemize}
\item GCN (Thomas N et al. 2017) \cite{DBLP:conf/iclr/KipfW17} learns the mean representation of neighboring node features, uses the mean as a weight, and aggregates the node features for the target node through multiple convolutions.
\item GAT (Petar Velickovic et al. 2019) \cite{DBLP:conf/iclr/VelickovicCCRLB18} enhances the GCN by introducing an attention mechanism that serves as the weight in the convolution process, with a particular emphasis on aggregating the features of neighboring nodes.

\item GCNII (Ming Chen et al. 2020) \cite{DBLP:conf/icml/ChenWHDL20} enhances the GCN model by incorporating initial residuals and constant mapping, along with arbitrary weights for multilayer convolution, effectively mitigating the over-smoothing problem.


\item ClusterGCN (Chiang et al. 2019) \cite{DBLP:conf/kdd/ChiangLSLBH19} uses a Metis strategy to partition the global graph into different subgraph partitions and randomly selects multiple partitioned subgraphs to participate in the training process.

\item GraphSAINT (Zeng et al. 2020) \cite{DBLP:conf/iclr/ZengZSKP20} reduces the problem of estimation bias in the training of traditional subgraph sampling algorithms by using a normalization method based on subgraph sampling.

\item LMC (Shi et al. 2023) \cite{DBLP:conf/iclr/ShiL023} applies the subgraph approach to the node classification task by employing a subgraph approach that retrieves information lost during forward and backward propagation in the neural network. Additionally, it incorporates nodes outside the subgraph into the GCN learning process.

\item ELPH (Chamberlain et al. 2023) \cite{chamberlain2023graph} applies the subgraph approach to the link prediction task by implicitly constructing subgraph sketches as message passing.

\item GCN-Ours, GAT-Ours, and GCNII-Ours methods utilize PromptGCN prompt templates to investigate the role of prompt in full-batch models.

\item ClusterGCN-Is, GraphSAINT-Is, and LMC-Is methods use the isolated prompt templates within each subgraph to investigate the role of prompts in subgraph sampling models. These methods are variants of PromptGCN, where each subgraph has its own exclusive prompt template, and the templates are independent of one another.

\item ClusterGCN-Ours, GraphSAINT-Ours, and LMC-Ours methods incorporate PromptGCN to explore the role of prompts in subgraph sampling models. These methods use a shared prompt template between subgraphs to transfer global information.
\end{itemize} 

\begin{table*}[!t]
\centering
\caption{Performance comparison among PromptGCN and the baselines on the node classification task at 3 layers.  The performance of PromptGCN is validated in three backbone models, respectively. \textbf{Bold} indicates the best results, while {\ul underlined} indicates the suboptimal results. ↑ denotes the higher score the better performance.}
\setlength{\tabcolsep}{2.6mm}
\begin{tabular}{clcccccccccc}
\toprule
\multicolumn{2}{l}{Models}                                                                                                                        & \multicolumn{2}{c}{\textbf{\begin{tabular}[c]{@{}c@{}}Amazon\\ CoBuyPhoto\end{tabular}}} & \multicolumn{2}{c}{\textbf{\begin{tabular}[c]{@{}c@{}}Amazon\\ CoBuyComputer\end{tabular}}} & \multicolumn{2}{c}{\textbf{Flickr}} & \multicolumn{2}{c}{\textbf{Reddit}} & \multicolumn{2}{c}{\textbf{Ogbn-products}} \\ \hline
\multicolumn{2}{l}{Metrics}                                                                                                                       & ACC↑                                        & F1↑                                        & ACC↑                                         & F1↑                                          & ACC↑             & F1↑              & ACC↑             & F1↑              & ACC↑                 & F1↑                 \\ \hline
\multicolumn{1}{c|}{\multirow{6}{*}{\begin{tabular}[c]{@{}c@{}}Full\\ batch\end{tabular}}}        & GCN \cite{DBLP:conf/iclr/KipfW17}             & 0.9385                                      & \textbf{0.9327}                            & \textbf{0.9116}                              & 0.9051                                       & 0.5226           & 0.2390           & \textbf{0.9442}  & \textbf{0.9152}  & \textbf{0.7560}      & \textbf{0.3484}     \\
\multicolumn{1}{c|}{}                                                                             & GCN-Ours                                      & \textbf{0.9411}                             & 0.9291                                     & 0.9102                                       & \textbf{0.9111}                              & \textbf{0.5314}  & \textbf{0.2438}  & 0.9440           & 0.9101           & 0.7405               & 0.3336              \\ \cline{2-12} 
\multicolumn{1}{c|}{}                                                                             & GAT \cite{DBLP:conf/iclr/VelickovicCCRLB18}   & \textbf{0.9509}                             & \textbf{0.9458}                            & 0.9193                                       & \textbf{0.9092}                              & 0.5162           & 0.2413           & 0.9401           & 0.9002           & \textbf{0.7552}      & 0.3077              \\
\multicolumn{1}{c|}{}                                                                             & GAT-Ours                                      & 0.9483                                      & 0.9414                                     & \textbf{0.9196}                              & 0.9091                                       & \textbf{0.5310}  & \textbf{0.2528}  & \textbf{0.9424}  & \textbf{0.9066}  & 0.7427               & \textbf{0.3121}     \\ \cline{2-12} 
\multicolumn{1}{c|}{}                                                                             & GCNII \cite{DBLP:conf/icml/ChenWHDL20}        & 0.9326                                      & 0.9289                                     & 0.9065                                       & 0.8979                                       & 0.4892           & 0.2292           & 0.9482           & 0.9131           & 0.7326               & 0.3217              \\
\multicolumn{1}{c|}{}                                                                             & GCNII-Ours                                    & \textbf{0.9359}                             & \textbf{0.9304}                            & \textbf{0.9127}                              & \textbf{0.9092}                              & \textbf{0.4945}  & \textbf{0.2445}  & \textbf{0.9527}  & \textbf{0.9303}  & \textbf{0.7433}      & \textbf{0.3665}     \\ \hline \hline
\multicolumn{1}{c|}{\multirow{9}{*}{\begin{tabular}[c]{@{}c@{}}Subgraph\\ Sampling\end{tabular}}} & ClusterGCN \cite{DBLP:conf/kdd/ChiangLSLBH19} & 0.9518                                      & 0.9430                                     & {\ul 0.9107}                                 & {\ul 0.9031}                                 & 0.4848           & 0.1619           & {\ul 0.9346}     & {\ul 0.9031}     & {\ul 0.7552}         & {\ul 0.3360}        \\
\multicolumn{1}{c|}{}                                                                             & ClusterGCN-Is                                 & {\ul 0.9529}                                & {\ul 0.9433}                               & 0.9065                                       & 0.8952                                       & {\ul 0.4923}     & {\ul 0.1805}     & 0.9323           & 0.8945           & 0.7365               & 0.3116              \\
\multicolumn{1}{c|}{}                                                                             & ClusterGCN-Ours                               & \textbf{0.9582}                             & \textbf{0.9497}                            & \textbf{0.9294}                              & \textbf{0.9233}                              & \textbf{0.5036}  & \textbf{0.2191}  & \textbf{0.9404}  & \textbf{0.9151}  & \textbf{0.7590}      & \textbf{0.3421}     \\ \cline{2-12} 
\multicolumn{1}{c|}{}                                                                             & GraphSAINT \cite{DBLP:conf/iclr/ZengZSKP20}   & 0.9735                                      & 0.9650                                     & 0.9315                                       & 0.9182                                       & 0.4760           & 0.1658           & {\ul 0.9639}     & {\ul 0.8992}     & {\ul 0.7271}         & {\ul 0.3336}        \\
\multicolumn{1}{c|}{}                                                                             & GraphSAINT-Is                                 & {\ul 0.9740}                                & {\ul 0.9681}                               & {\ul 0.9325}                                 & {\ul 0.9205}                                 & {\ul 0.4856}     & {\ul 0.1751}     & 0.9622           & 0.8873           & 0.6986               & 0.3163              \\
\multicolumn{1}{c|}{}                                                                             & GraphSAINT-Ours                               & \textbf{0.9763}                             & \textbf{0.9716}                            & \textbf{0.9373}                              & \textbf{0.9291}                              & \textbf{0.5021}  & \textbf{0.2349}  & \textbf{0.9688}  & \textbf{0.9116}  & \textbf{0.7324}      & \textbf{0.3375}     \\ \cline{2-12} 
\multicolumn{1}{c|}{}                                                                             & LMC \cite{DBLP:conf/iclr/ShiL023}             & {\ul 0.9000}                                & 0.8910                                     & {\ul 0.8400}                                 & {\ul 0.8372}                                 & \textbf{0.5392}  & {\ul 0.2936}     & \textbf{0.9544}  & 0.9342           & 0.7409               & {\ul 0.3519}        \\
\multicolumn{1}{c|}{}                                                                             & LMC-Is                                        & 0.8875                                      & {\ul 0.8933}                               & 0.7750                                       & 0.7731                                       & 0.5361           & 0.2909           & 0.9532           & {\ul 0.9344}     & {\ul 0.7418}         & \textbf{0.3593}     \\
\multicolumn{1}{c|}{}                                                                             & LMC-Ours                                      & \textbf{0.9187}                             & \textbf{0.9090}                            & \textbf{0.8500}                              & \textbf{0.8493}                              & {\ul 0.5387}     & \textbf{0.2966}  & {\ul 0.9542}     & \textbf{0.9352}  & \textbf{0.7469}      & 0.3434              \\ \bottomrule
\end{tabular}\label{TT}
\end{table*}

\begin{figure*}[!t]
  \centering
  \includegraphics[width=1\linewidth]{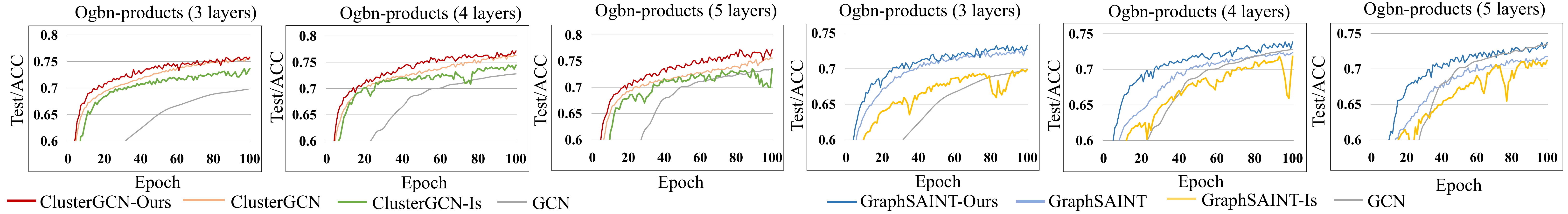}
  \caption{Performance comparison of promptGCN with baselines on different layers. PromptGCN achieves the highest ACC across different layers.}
  \label{Fn1234}
\end{figure*}

\begin{table}[!t]
\centering
\caption{Performance comparison between PromptGCN and the baseline on the link prediction task at 3 layers.  The performance of PromptGCN is validated in three backbone models, respectively. \textbf{Bold} indicates the best results, while {\ul underlined} indicates the suboptimal results. ↑ denotes the higher score the better performance.}
\setlength{\tabcolsep}{2.9mm}
\begin{tabular}{clcc}
\toprule
\multicolumn{2}{l}{Models}                                                                                                                        & \textbf{Ogbl-citation2} & \textbf{Ogbl-collab} \\ \hline
\multicolumn{2}{l}{Metrics}                                                                                                                       & MRR↑                    & Hits@100↑            \\ \hline
\multicolumn{1}{c|}{\multirow{6}{*}{\begin{tabular}[c]{@{}c@{}}Full\\ Batch\end{tabular}}}        & GCN \cite{DBLP:conf/iclr/KipfW17}             & 0.8432                  & 0.4557               \\
\multicolumn{1}{c|}{}                                                                             & GCN-Ours                                      & OOM                     & \textbf{0.4795}      \\ \cline{2-4} 
\multicolumn{1}{c|}{}                                                                             & GAT \cite{DBLP:conf/iclr/VelickovicCCRLB18}   & 0.8434                  & \textbf{0.4322}      \\
\multicolumn{1}{c|}{}                                                                             & GAT-Ours                                      & OOM                     & 0.4228               \\ \cline{2-4} 
\multicolumn{1}{c|}{}                                                                             & GCNII \cite{DBLP:conf/icml/ChenWHDL20}        & 0.8435                  & \textbf{0.4664}      \\
\multicolumn{1}{c|}{}                                                                             & GCNII-Ours                                    & OOM                     & 0.4548               \\ \hline\hline
\multicolumn{1}{c|}{\multirow{9}{*}{\begin{tabular}[c]{@{}c@{}}Subgraph\\ Sampling\end{tabular}}} & ClusterGCN \cite{DBLP:conf/kdd/ChiangLSLBH19} & {\ul 0.8076}            & {\ul 0.4889}         \\
\multicolumn{1}{c|}{}                                                                             & ClusterGCN-Is                                 & 0.7323                  & 0.3779               \\
\multicolumn{1}{c|}{}                                                                             & ClusterGCN-Ours                               & \textbf{0.8118}         & \textbf{0.4988}      \\ \cline{2-4} 
\multicolumn{1}{c|}{}                                                                             & GraphSAINT \cite{DBLP:conf/iclr/ZengZSKP20}   & {\ul 0.7933}            & {\ul 0.4745}         \\
\multicolumn{1}{c|}{}                                                                             & GraphSAINT-Is                                 & 0.7765                  & 0.4642               \\
\multicolumn{1}{c|}{}                                                                             & GraphSAINT-Ours                               & \textbf{0.8023}         & \textbf{0.5289}      \\ \cline{2-4} 
\multicolumn{1}{c|}{}                                                                             & ELPH \cite{chamberlain2023graph}              & \textbf{0.8756}         & {\ul 0.7094}               \\
\multicolumn{1}{c|}{}                                                                             & ELPH-Is                                       & 0.8442                       & 0.6878                   \\
\multicolumn{1}{c|}{}                                                                             & ELPH-Ours                                     & {\ul 0.8642}                  & \textbf{0.7114}      \\ \hline
\end{tabular}\label{TTTTTT}
\end{table}

\textbf{Evaluation indicators.} This paper evaluates the PromptGCN model using several classical evaluation metrics. 
As shown in Table \ref{2222}. 1) For the node classification task, \textit{Accuracy} and marco-\textit{F1} are used as an evaluation metric on all datasets. 2) For the link prediction task, we adopt the same evaluation metrics as \cite{chamberlain2023graph}. The Obgl-citation2 dataset emphasizes the precise ordering of predictions, and therefore the \textit{MRR} metric is used. In contrast, the Obgl-collab dataset focuses on the coverage of predictions, and therefore the \textit{Hits@100} metric is adopted.

\textbf{Hyperparameters.} For all the experimental models, this paper sets the learning rate as 0.001, dropout as 0.5, and weight decay as 0.0005. In addition, we set the value of the hyperparameter $\gamma$ to 0.1. On different sampling methods, we adopt the same batch size and hidden dimension. In the main experiment, the number of prompts was aligned with the number of labels. Experiments are conducted on a machine with a NVIDIA A100 GPU (80 GB memory).

\begin{table*}[!t]
\centering
\caption{Comparison of memory consumption between PromptGCN and baselines at different numbers of layers (measured in * $10^2$ MB). \textbf{Bold}  indicates where memory consumption is lower than the full-batch method. The lower metric the smaller memory consumption.}
\setlength{\tabcolsep}{5.1mm}
\begin{tabular}{llllccclclc}
\toprule
\textbf{Layer}                                & \multicolumn{3}{l}{\textbf{Models}}                               & \textbf{\begin{tabular}[c]{@{}c@{}}Amazon\\ CoBuyPhoto\end{tabular}} & \textbf{\begin{tabular}[c]{@{}c@{}}Amazon\\ CoBuyComputer\end{tabular}} & \multicolumn{2}{c}{\textbf{Flickr}} & \multicolumn{2}{c}{\textbf{Reddit}} & \textbf{Ogbn-products} \\ \hline
\multicolumn{1}{l|}{\multirow{7}{*}{3-layer}} & \multicolumn{3}{l}{GCN \cite{DBLP:conf/iclr/KipfW17}}             & 0.99                                                                 & 1.81                                                                    & \multicolumn{2}{c}{9.41}            & \multicolumn{2}{c}{26}              & 182                    \\
\multicolumn{1}{l|}{}                         & \multicolumn{3}{l}{ClusterGCN \cite{DBLP:conf/kdd/ChiangLSLBH19}} & 0.25                                                                 & 0.38                                                                    & \multicolumn{2}{c}{1.86}            & \multicolumn{2}{c}{5.15}            & 27.78                  \\
\multicolumn{1}{l|}{}                         & \multicolumn{3}{l}{ClusterGCN-Is}                                 & 0.67                                                                 & 1.11                                                                    & \multicolumn{2}{c}{4.08}            & \multicolumn{2}{c}{12.13}           & 31.23                  \\
\multicolumn{1}{l|}{}                         & \multicolumn{3}{l}{ClusterGCN-Ours}                               & \textbf{0.25}                                                        & \textbf{1.12}                                                           & \multicolumn{2}{c}{\textbf{4.27}}   & \multicolumn{2}{c}{\textbf{8.83}}   & \textbf{37.42}         \\
\multicolumn{1}{l|}{}                         & \multicolumn{3}{l}{GraphSAINT \cite{DBLP:conf/iclr/ZengZSKP20}}   & 0.45                                                                 & 0.52                                                                    & \multicolumn{2}{c}{0.58}            & \multicolumn{2}{c}{0.97}            & 5.01                   \\
\multicolumn{1}{l|}{}                         & \multicolumn{3}{l}{GraphSAINT-Is}                                 & 1.32                                                                 & 1.49                                                                    & \multicolumn{2}{c}{1.29}            & \multicolumn{2}{c}{2.03}            & 5.97                   \\
\multicolumn{1}{l|}{}                         & \multicolumn{3}{l}{GraphSAINT-Ours}                               & \textbf{1.13}                                                        & \textbf{1.15}                                                           & \multicolumn{2}{c}{\textbf{1.36}}   & \multicolumn{2}{c}{\textbf{1.46}}   & \textbf{7.69}          \\ \hline\hline
\multicolumn{1}{l|}{\multirow{7}{*}{4-layer}} & \multicolumn{3}{l}{GCN \cite{DBLP:conf/iclr/KipfW17}}             & 1.18                                                                 & 2.15                                                                    & \multicolumn{2}{c}{11.1}            & \multicolumn{2}{c}{32.23}           & 243                    \\
\multicolumn{1}{l|}{}                         & \multicolumn{3}{l}{ClusterGCN \cite{DBLP:conf/kdd/ChiangLSLBH19}} & 0.33                                                                 & 0.51                                                                    & \multicolumn{2}{c}{2.49}            & \multicolumn{2}{c}{6.72}            & 34.57                  \\
\multicolumn{1}{l|}{}                         & \multicolumn{3}{l}{ClusterGCN-Is}                                 & 0.69                                                                 & 1.14                                                                    & \multicolumn{2}{c}{4.11}            & \multicolumn{2}{c}{12.83}           & 39.44                  \\
\multicolumn{1}{l|}{}                         & \multicolumn{3}{l}{ClusterGCN-Ours}                               & \textbf{0.69}                                                        & \textbf{1.13}                                                           & \multicolumn{2}{c}{\textbf{4.91}}   & \multicolumn{2}{c}{\textbf{13.22}}  & \textbf{45.43}         \\
\multicolumn{1}{l|}{}                         & \multicolumn{3}{l}{GraphSAINT \cite{DBLP:conf/iclr/ZengZSKP20}}   & 0.59                                                                 & 0.68                                                                    & \multicolumn{2}{c}{0.78}            & \multicolumn{2}{c}{1.25}            & 6.63                   \\
\multicolumn{1}{l|}{}                         & \multicolumn{3}{l}{GraphSAINT-Is}                                 & 1.28                                                                 & 1.51                                                                    & \multicolumn{2}{c}{1.33}            & \multicolumn{2}{c}{2.13}            & 7.60                   \\
\multicolumn{1}{l|}{}                         & \multicolumn{3}{l}{GraphSAINT-Ours}                               & 1.29                                                                 & \textbf{1.51}                                                           & \multicolumn{2}{c}{\textbf{1.43}}   & \multicolumn{2}{c}{\textbf{1.68}}   & \textbf{9.39}          \\ \hline\hline
\multicolumn{1}{l|}{\multirow{7}{*}{5-layer}} & \multicolumn{3}{l}{GCN \cite{DBLP:conf/iclr/KipfW17}}             & 1.37                                                                 & 2.49                                                                    & \multicolumn{2}{c}{14.1}            & \multicolumn{2}{c}{37.21}           & 292                    \\
\multicolumn{1}{l|}{}                         & \multicolumn{3}{l}{ClusterGCN \cite{DBLP:conf/kdd/ChiangLSLBH19}} & 0.40                                                                 & 0.61                                                                    & \multicolumn{2}{c}{3.13}            & \multicolumn{2}{c}{8.33}            & 42.81                  \\
\multicolumn{1}{l|}{}                         & \multicolumn{3}{l}{ClusterGCN-Is}                                 & 0.71                                                                 & 1.15                                                                    & \multicolumn{2}{c}{4.60}            & \multicolumn{2}{c}{12.92}           & 47.71                  \\
\multicolumn{1}{l|}{}                         & \multicolumn{3}{l}{ClusterGCN-Ours}                               & \textbf{0.78}                                                        & \textbf{1.22}                                                           & \multicolumn{2}{c}{\textbf{5.55}}   & \multicolumn{2}{c}{\textbf{15.25}}  & \textbf{53.21}         \\
\multicolumn{1}{l|}{}                         & \multicolumn{3}{l}{GraphSAINT \cite{DBLP:conf/iclr/ZengZSKP20}}   & 0.72                                                                 & 0.84                                                                    & \multicolumn{2}{c}{0.99}            & \multicolumn{2}{c}{1.53}            & 8.24                   \\
\multicolumn{1}{l|}{}                         & \multicolumn{3}{l}{GraphSAINT-Is}                                 & 1.30                                                                 & 1.53                                                                    & \multicolumn{2}{c}{1.46}            & \multicolumn{2}{c}{2.15}            & 9.23                   \\
\multicolumn{1}{l|}{}                         & \multicolumn{3}{l}{GraphSAINT-Ours}                               & 1.39                                                                 & \textbf{1.67}                                                           & \multicolumn{2}{c}{\textbf{1.76}}   & \multicolumn{2}{c}{\textbf{1.88}}   & \textbf{10.91}         \\ \bottomrule
\end{tabular}
\label{T5}
\end{table*}

\subsection{Prediction Performance Comparison (RQ1)}\label{5.2}
This section compares the performance of PromptGCN with baselines on seven large-scale datasets. Table \ref{TT} and Table \ref{TTTTTT} show that the performance comparison among PromptGCN and the baselines on the \textit{node classification} and \textit{link prediction} tasks at 3 layers,  respectively. The experimental results are statistical results of multiple experiments. 
We apply PromptGCN to two representative subgraph sampling methods (i.e., ClusterGCN \cite{DBLP:conf/kdd/ChiangLSLBH19} and GraphSAINT \cite{DBLP:conf/iclr/ZengZSKP20}) and two latest methods (i.e., LMC \cite{DBLP:conf/iclr/ShiL023} and ELPH \cite{chamberlain2023graph}). From the experimental results, we can draw the following conclusions:

(1) PromptGCN is applied to node classification (Table \ref{TT}) and link prediction (Table \ref{TTTTTT}) tasks, respectively. Experiments show that PromptGCN finally resembles full-batch performance on all datasets. 

\begin{figure*}[!t]
  \centering
  \includegraphics[width=1\linewidth]{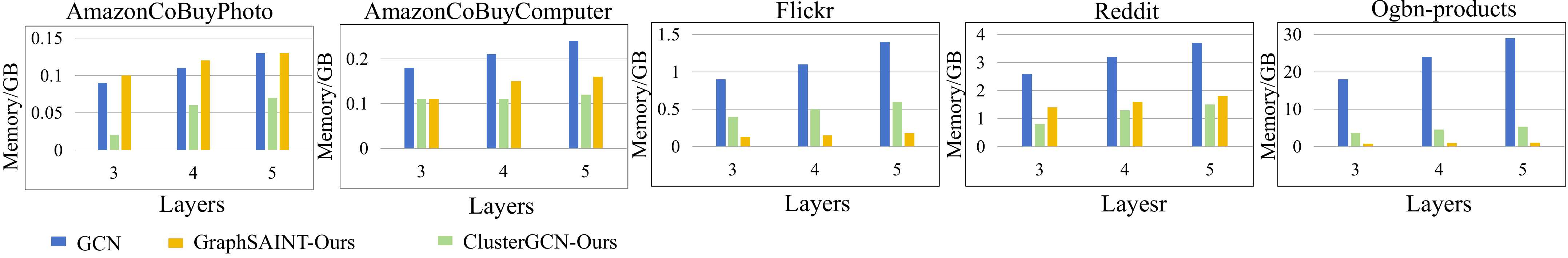}
  \caption{Memory consumption of PromptGCN and compared models with different numbers of layers.}
  \label{Fn44}
\end{figure*}

(2) Table \ref{TT} presents the overall performance of PromptGCN on the node classification task. Compared to the three backbone subgraph sampling models, PromptGCN achieves competitive metrics across almost all datasets. For instance, on the Flickr dataset, PromptGCN improves performance by 5.48\%. Additionally, we selected the largest dataset, Ogbn-products, for visualization experiments. As shown in Fig. \ref{Fn1234}, experiments across different layers reveal that PromptGCN achieves the highest test accuracy. 

(3) To evaluate the performance of PromptGCN on different downstream tasks, we applied it to a link prediction task. Table \ref{TTTTTT} presents the overall performance of PromptGCN for this task. On the Ogbl-collab dataset, PromptGCN improves the backbone performance by $2.02\%$. The strong performance of PromptGCN across various tasks demonstrates that our method is adaptable to different backbone models and downstream tasks. Experimental results on two tasks show that PromptGCN performs less impressively with the latest backbone models. This suggests that these advanced neural network models are more complex and require more sophisticated and well-designed prompts.

(4)  Additionally, when applied to full-batch models, PromptGCN does not consistently improve metrics across all experiments. For example, on the Ogbn-products dataset, GCNII-Ours improves performance by 1.46\% and 13.92\% on both metrics, whereas GCN-Ours fails to match the performance of the original model. The reason is that when training a full-batch model on the entire graph, PromptGCN cannot effectively bridge the gaps between subgraphs, a challenge unique to subgraph sampling algorithms. It’s important to note that due to the high memory consumption of the full-batch models, using PromptGCN with it can result in an OOM error.

(5) ``-Is'' denotes the isolated prompt assigned to each subgraph. The isolated prompt fails to achieve the superior performance of PromptGCN. The reason is that isolated prompt assigns independent prompts to each subgraph, disrupting the role of prompts as a bridge between subgraphs, and making it challenging to expand the subgraph receptive field or transfer global information. In contrast, the shared prompt significantly strengthens the connection between subgraphs by capturing global information during training.

In summary, PromptGCN can effectively bridge the gap among subgraphs and can be generalized to different subgraph sampling methods as well as different downstream tasks to achieve competitive performance.
\begin{figure*}[!t]
  \centering
  \includegraphics[width=1\linewidth]{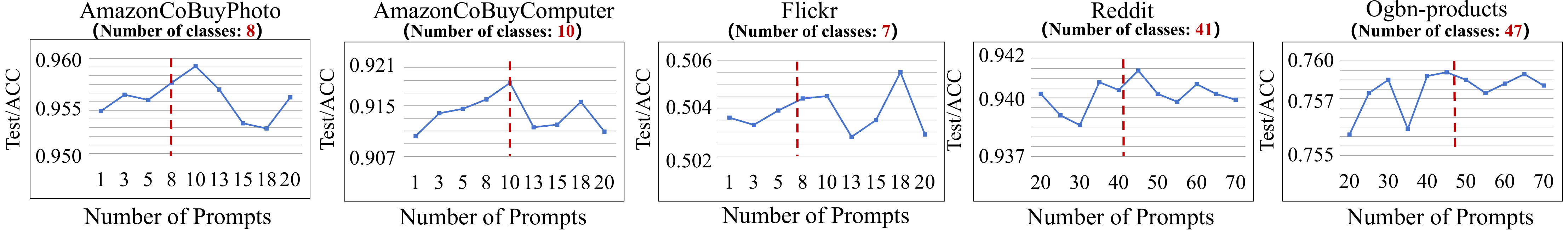}
  \caption{Impact of different number of prompts on test accuracy, and the red dotted line represents the number of classes. The prompt embeddings are positioned close to their corresponding classification clusters, indicating that aligning the number of prompts with the number of classifications optimizes model performance.}
  \label{Fn011}
\end{figure*}

\subsection{Memory Consumption Comparison (RQ2)}
The deeper aggregation layers of full-batch GCN significantly increase the memory consumption. In Table \ref{T5}, we compare the memory consumption of PromptGCN, GraphGCN, GraphSAINT and GCN across different layers. From the experimental results, we can draw the following conclusions: 

(1) In terms of the number of aggregation layers, PromptGCN generally uses less memory than full-batch GCN, with the difference increasing as the number of layers grows, reaching up to 8 times less. Fig. \ref{Fn44} visually demonstrates that PromptGCN's memory consumption is significantly lower than that of full-batch GCN. 

(2) In terms of dataset scale, as the size of the data increases, PromptGCN exhibits significantly lower memory consumption than GCN on large-scale datasets (e.g., Ogbn-products), with reductions of up to 28 times.

(3) Although the memory consumption of PromptGCN is slightly larger than that of the backbone models (ClusterGCN and GraphSAINT), it remains within acceptable limits. Moreover, Table \ref{TT} illustrated that PromptGCN finally resembles full-batch performance and outperforms subgraph sampling models. 

To summarize, PromptGCN has a lower memory consumption than a full-batch while achieving higher accuracy.

\begin{figure}[!t]
\centering
\subfigure[AmazonCoBuyPhoto]{\includegraphics[height=3cm,width=4cm]{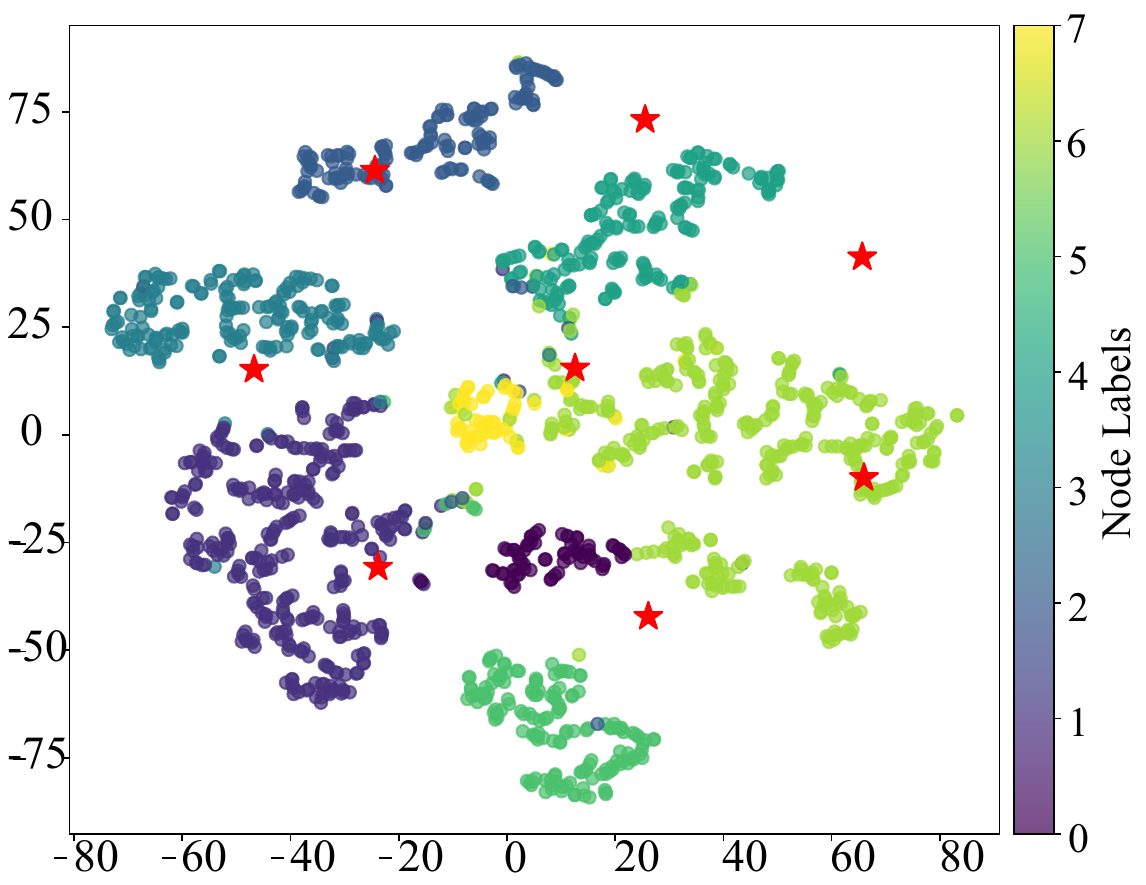}}
\subfigure[AmazonCoBuyComputer]{\includegraphics[height=3cm,width=4cm]{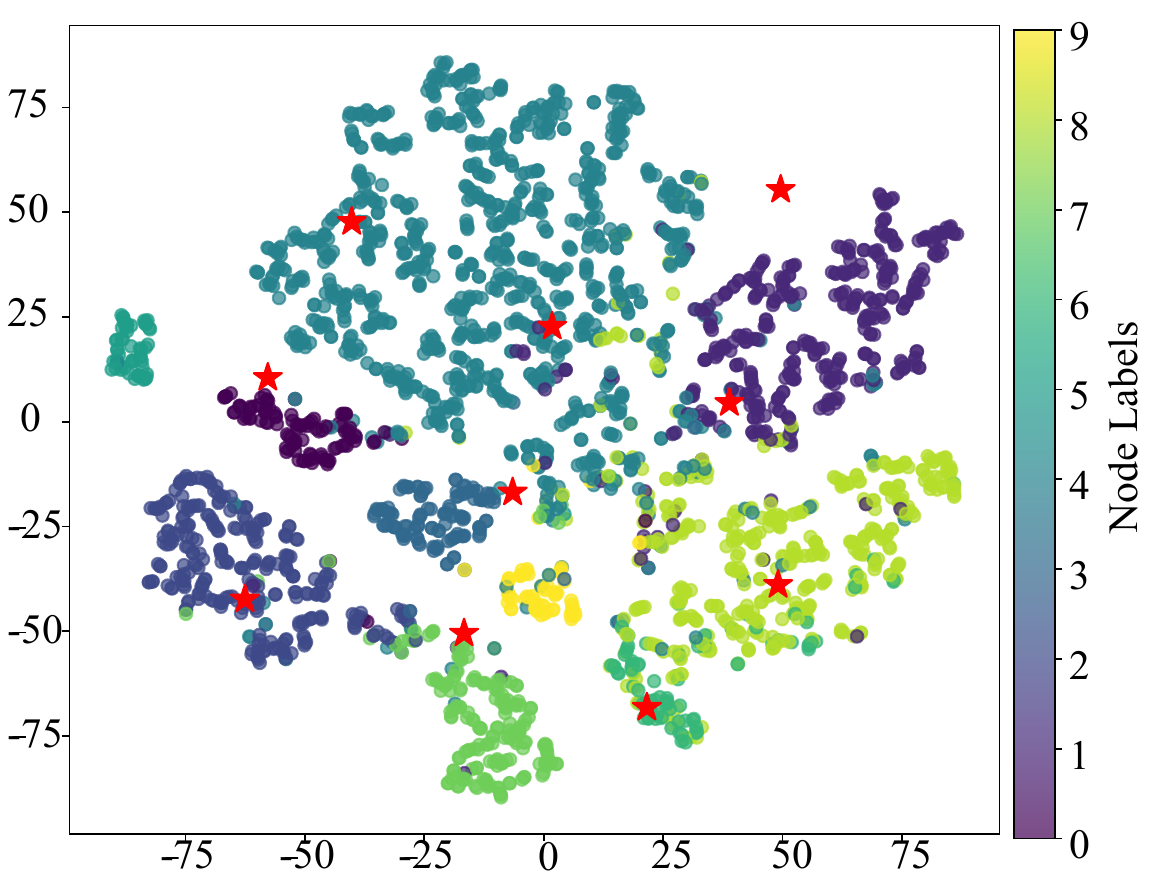}}
\caption{Visualization of nodes and prompts on (a) AmazonCoBuyPhoto and (b) AmazonCoBuyComputer datasets.}
\label{F4}
\end{figure}

\begin{figure*}[!h]
\centering
\subfigure[CulsterGCN-Ours]{
    \includegraphics[width=1\linewidth]{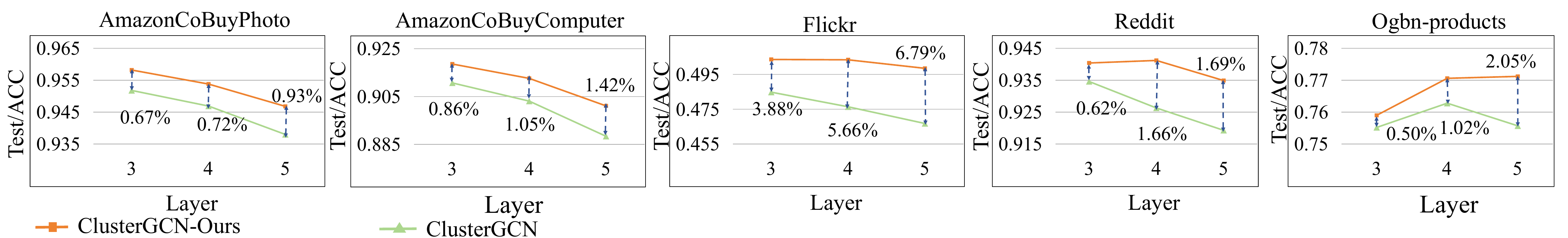}
    \label{Fn(a)}
}
\hspace{0.5cm}
\subfigure[GraphSAINT-Ours]{
    \includegraphics[width=1\linewidth]{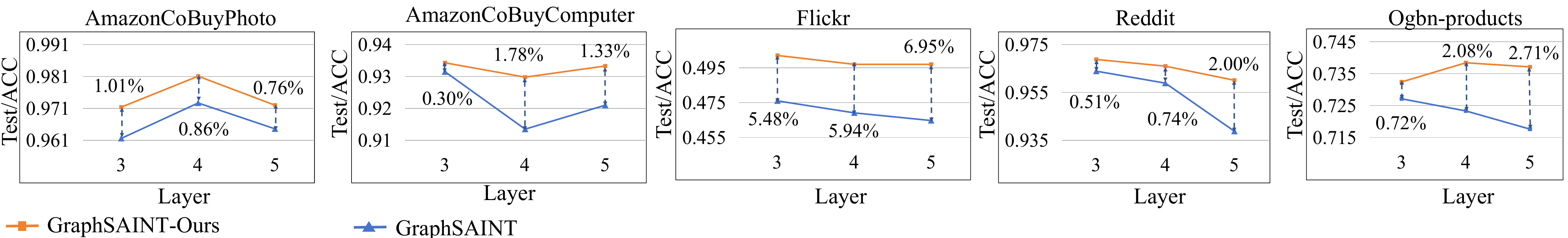}
    \label{Fn(b)}
}
\caption{Deeper layers of PromptGCN on ClusterGCN (a) and GraphSAINT (b). PromptGCN effectively mitigates the over-smoothing issue in deeper GCN models by integrating global information through prompt parameters across subgraphs.}
\label{Fn}
\end{figure*}

\begin{table*}[!t]
\centering
\caption{Effect of different attachment modes of ClusterGCN-Ours on test accuracy. Integrated
prompt with node features in different attachment methods, including Concatenate, Point-wise Addition, Point-wise Multiplication, and Point-wise Weighted Addition. \textbf{Bold} indicates the best results. Use ACC as the evaluation metric, and the higher score the better performance.}
\setlength{\tabcolsep}{5.2mm}
\begin{tabular}{llccclc}
\toprule
\multicolumn{2}{l}{\textbf{\begin{tabular}[c]{@{}l@{}}Attachment patterns\end{tabular}}} & \textbf{\begin{tabular}[c]{@{}c@{}}Amazon\\ CoBuyPhoto\end{tabular}} & \textbf{\begin{tabular}[c]{@{}c@{}}Amazon\\ CoBuyComputer\end{tabular}} & \textbf{Flickr} & \multicolumn{1}{c}{\textbf{Reddit}} & \textbf{Ogbn-products} \\ \hline
\multicolumn{2}{l|}{Concatenate}                                                                 & \textbf{0.9582}                                                      & \textbf{0.9294}                                                         & \textbf{0.5036} & 0.9404                              & \textbf{0.7590}        \\ \hline
\multicolumn{2}{l|}{Point-wise Addition}                                                                    & 0.9535                                                               & 0.9223                                                                  & 0.5012          & 0.9376                              & 0.7482                 \\ \hline
\multicolumn{2}{l|}{Point-wise Multiplication}                                                                   & 0.9524                                                               & 0.9114                                                                  & 0.4961          & 0.9303                              & 0.7522                 \\ \hline
\multicolumn{2}{l|}{Point-wise Weighted Addition ($\alpha$=0.2)}                                            & 0.9549                                                               & 0.9206                                                                  & 0.5036          & 0.9384                              & 0.7500                 \\ \hline
\multicolumn{2}{l|}{Point-wise Weighted Addition ($\alpha$=0.4)}                                            & 0.9535                                                               & 0.9173                                                                  & 0.5011          & \textbf{0.9411}                     & 0.7462                 \\ \hline
\multicolumn{2}{l|}{Point-wise Weighted Addition ($\alpha$=0.6)}                                            & 0.9522                                                               & 0.9276                                                                  & 0.5036          & 0.9399                              & 0.7546                 \\ \hline
\multicolumn{2}{l|}{Point-wise Weighted Addition ($\alpha$=0.8)}                                            & 0.9575                                                               & 0.9247                                                                  & 0.5028          & 0.9408                              & 0.7578                 \\ \bottomrule
\end{tabular}\label{T8}
\end{table*}

\begin{table*}[!t]
\centering
\caption{Test Accuracy of ClusterGCN and ClusterGCN-Ours at different number of subgraphs, and \textbf{bold} indicates the best results. Use ACC as the evaluation metric, and the higher score the better performance.}
\setlength{\tabcolsep}{5.2mm}
\begin{tabular}{clllccclclc}
\toprule
\multicolumn{1}{l}{\textbf{\begin{tabular}[c]{@{}l@{}}Number of \\ subgraphs\end{tabular}}} & \multicolumn{3}{l}{\textbf{Models}} & \textbf{\begin{tabular}[c]{@{}c@{}}Amazon\\ CoBuyPhoto\end{tabular}} & \textbf{\begin{tabular}[c]{@{}c@{}}Amazon\\ CoBuyComputer\end{tabular}} & \multicolumn{2}{c}{\textbf{Flickr}} & \multicolumn{2}{c}{\textbf{Reddit}} & \textbf{Ogbn-products} \\ \hline
\multicolumn{1}{c|}{\multirow{2}{*}{100}}                                                   & \multicolumn{3}{l}{ClusterGCN \cite{DBLP:conf/kdd/ChiangLSLBH19}}      & 0.9464                                                               & 0.8960                                                                  & \multicolumn{2}{c}{0.4838}          & \multicolumn{2}{c}{0.9303}          & 0.7658                 \\
\multicolumn{1}{c|}{}                                                                       & \multicolumn{3}{l}{ClusterGCN-Ours} & \textbf{0.9503}                                                      & \textbf{0.9076}                                                         & \multicolumn{2}{c}{\textbf{0.4924}} & \multicolumn{2}{c}{\textbf{0.9327}} & \textbf{0.7707}        \\ \hline
\multicolumn{1}{c|}{\multirow{2}{*}{20}}                                                    & \multicolumn{3}{l}{ClusterGCN \cite{DBLP:conf/kdd/ChiangLSLBH19}}      & 0.9507                                                               & 0.9102                                                                  & \multicolumn{2}{c}{0.4864}          & \multicolumn{2}{c}{0.9365}          & 0.7683                 \\
\multicolumn{1}{c|}{}                                                                       & \multicolumn{3}{l}{ClusterGCN-Ours} & \textbf{0.9533}                                                      & \textbf{0.9160}                                                         & \multicolumn{2}{c}{\textbf{0.5026}} & \multicolumn{2}{c}{\textbf{0.9384}} & \textbf{0.7629}        \\ \hline
\multicolumn{1}{c|}{\multirow{2}{*}{10}}                                                    & \multicolumn{3}{l}{ClusterGCN \cite{DBLP:conf/kdd/ChiangLSLBH19}}      & 0.9481                                                               & 0.9113                                                                  & \multicolumn{2}{c}{0.4869}          & \multicolumn{2}{c}{0.9352}          & 0.7552                 \\
\multicolumn{1}{c|}{}                                                                       & \multicolumn{3}{l}{ClusterGCN-Ours} & \textbf{0.9604}                                                      & \textbf{0.9184}                                                         & \multicolumn{2}{c}{\textbf{0.5018}} & \multicolumn{2}{c}{\textbf{0.9392}} & \textbf{0.7590}        \\ \hline
\multicolumn{1}{c|}{\multirow{2}{*}{5}}                                                     & \multicolumn{3}{l}{ClusterGCN \cite{DBLP:conf/kdd/ChiangLSLBH19}}      & 0.9518                                                               & 0.9107                                                                  & \multicolumn{2}{c}{0.4848}          & \multicolumn{2}{c}{0.9346}          & 0.7455                 \\
\multicolumn{1}{c|}{}                                                                       & \multicolumn{3}{l}{ClusterGCN-Ours} & \textbf{0.9582}                                                      & \textbf{0.9294}                                                         & \multicolumn{2}{c}{\textbf{0.5036}} & \multicolumn{2}{c}{\textbf{0.9404}} & \textbf{0.7530}        \\ \bottomrule
\end{tabular}\label{T4}
\end{table*}



\subsection{The Analysis of Hyperparameters (RQ3)}\label{5.6}
\textbf{Impact of Different Number of Prompts.} Fig. \ref{Fn011} illustrates the effect of different numbers of prompts on model performance. In most cases, performance is optimized when the number of prompts matches the number of classifications. Inspired by this observation, it’s crucial to explore the placement of prompt embeddings within the subgraph. Fig. \ref{F4} presents a visualization of node embeddings and prompt embeddings. Colored circles represent nodes, while red pentagrams mark the locations of prompt embeddings in the subgraph, with different colors denoting different categories.
In conclusion, the prompt embeddings are positioned close to their corresponding classification clusters, indicating that aligning the number of prompts with the number of classifications optimizes model performance.

\textbf{Impact of Different Number of Layers.} We explore the impact of deeper PromptGCN in our experiments. The prediction trend of using PromptGCN on two classical models, ClusterGCN and GraphSAINT, is illustrated in Fig. \ref{Fn}. In most cases, the disparity between the PromptGCN and the original model increases as the number of layers deepens, with this difference reflected in the predictive performance of the model on downstream tasks. 
For instance, on the Flickr dataset, PromptGCN boosts performance by 5.48\% at 3 layers, while it increases to 6.95\% at 5 layers.
With an increase in the number of aggregation layers, the model acquires more information about neighboring nodes, leading to the phenomenon of over-smoothing. In other words, an excessive number of learned node representations accelerates the homogenization of nodes, making it challenging to distinguish node differences, thus resulting in degraded model performance. 
In contrast, PromptGCN learns global information not by aggregating neighboring nodes but by supplementing global information through prompt parameters across different subgraphs. This approach avoids placing additional burden on the model, even as the number of layers increases.

\textbf{Impact of Different Prompt Attachment Methods.} We integrated Prompts with node features using various attachment methods, including concatenate, point-wise addition, point-wise multiplication, and point-wise weighted addition. Table \ref{T8} presents the experimental results on the five datasets, indicating that: 1) The concatenate scheme achieves optimal results because it maximizes the retention of the independence between prompts and node embedding features. 2) Other attachment methods also contribute to model performance. 
We find that using the concatenate operation to attach prompt
embeddings to node features maximize the delivery of global
information. Therefore, in this paper, the concatenate operation
is used as the main attachment method.

\textbf{Impact of Different Numbers of Subgraphs.} To evaluate the performance of PromptGCN with varying numbers of subgraphs, we conducted experiments by adjusting the number of subgraphs in the ClusterGCN model. Table \ref{T4} presents the ACC across five datasets. The results show that: 1) The ClusterGCN results confirm our hypothesis from the introduction that the subgraph receptive field is negatively correlated with the number of subgraphs—i.e., as more subgraphs are partitioned, more global information is lost, which significantly impacts the prediction performance of downstream tasks. 2) PromptGCN consistently outperforms ClusterGCN across all subgraph configurations, demonstrating that the prompting strategy effectively expands the subgraph receptive field and captures global information.


\section{Related Work}
\label{related}
\begin{figure*}[!t]
  \centering
  \includegraphics[width=0.85\linewidth]{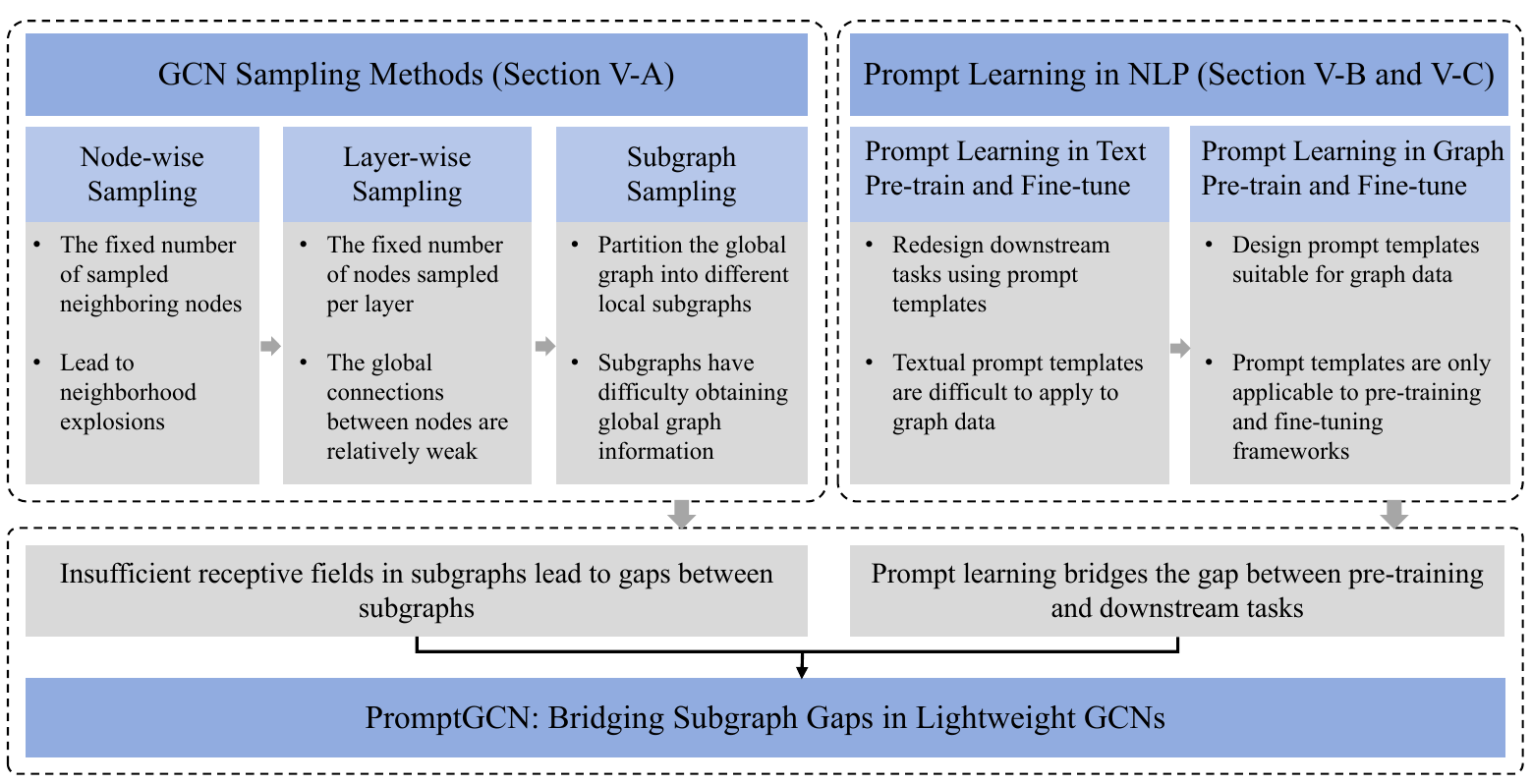}
  \caption{Summary of related work. This section contains two parts: GCN sampling methods and prompt learning in MLP.} 
  \label{R11}
\end{figure*}

The section presents related work from two perspectives, including GCN Sampling (\ref{R1}), Prompt Learning in Text Pre-train and Fine-tune (\ref{R2}), and Prompt Learning in Graph Pre-train and Fine-tune (\ref{R3}). The research framework for this section is illustrated in Fig. \ref{R11}.
\subsection{GCN Sampling}\label{R1}
GCNs are widely used in graph-based applications, such as social networks and recommender systems \cite{chen2023improving}, \cite{gao2023cirs}, to learn the feature representation of target nodes by aggregating information from multiple layers of neighboring nodes during full-batch training. Nevertheless, this approach increases additional memory consumption \cite{DBLP:conf/asplos/YangZD023}.
To effectively reduce memory consumption, GCN sampling methods use mini-batch training at various granularities, serving as a crucial technique for lightweight GCNs.

\textbf{Node-wise Sampling Methods.} The node-wise sampling methods \cite{DBLP:conf/nips/HamiltonYL17}, \cite{DBLP:conf/icml/ChenZS18} sample neighbors for each node, eliminating the need for GCNs to process the entire graph, thereby reducing the memory consumption generated during the training process. 
GraphSAGE \cite{DBLP:conf/nips/HamiltonYL17} randomly samples a fixed number of its first-order neighbors to reduce the memory consumption. Unlike GraphSAGE, which optimizes the sampling strategy, \cite{DBLP:conf/icml/ChenZS18} proves the convergence of its sampling algorithm. 
The above methods sample nodes with random probability. \cite{DBLP:journals/nn/ZhaoGYH23} proposes a learnable sampling method that samples nodes with an unfixed probability.
\cite{DBLP:conf/mlsys/WanLLKL22} introduce the node-wise sampling method to reduce the number of redundant boundary nodes per partitioned subgraph and reduce GCN communication costs. However, the number of nodes sampled grows exponentially with the number of GCN layers, leading to the node explosion problem.

\textbf{Layer-wise Sampling Methods.} The layer-wise sampling methods \cite{DBLP:conf/iclr/ChenMX18}, \cite{DBLP:conf/nips/ZouHWJSG19}, \cite{DBLP:conf/nips/Huang0RH18} sample a certain number of nodes for each GCN layer to alleviate the node explosion problem. 
\cite{DBLP:conf/iclr/ChenMX18} and \cite{DBLP:conf/nips/ZouHWJSG19} sample a fixed number of nodes in each layer, but \cite{DBLP:conf/nips/LiuW00YS020} and \cite{DBLP:journals/tmlr/ChenXHJYZ23} believe that sampling an incremental number of nodes in each layer produces better results.
\cite{DBLP:conf/nips/BalinC23} combines the advantages of layer-wise sampling and node-wise sampling to sample the minimum number of vertices needed for each layer.
However, these methods only consider node associations between neighboring GCN layers; information loss from unsampled nodes in each layer will result in a reduction in GCN accuracy.

\textbf{Subgraph Sampling Methods}. The subgraph sampling methods \cite{DBLP:conf/kdd/ChiangLSLBH19}, \cite{DBLP:conf/iclr/ZengZSKP20}, \cite{DBLP:conf/iclr/ShiL023} reduces the memory consumption of GCN while ensuring that all nodes are sampled. 
ClusterGCN \cite{DBLP:conf/kdd/ChiangLSLBH19} uses a METIS \cite{DBLP:journals/siamsc/KarypisK98} strategy to partition the global graph into different subgraphs and randomly selects multiple partitioned subgraphs to participate in the training process. 
GraphSAINT \cite{DBLP:conf/kdd/ChiangLSLBH19} reduces the problem of estimation bias in the training of traditional subgraph sampling algorithms by applying a normalization method based on subgraph sampling. 
\cite{DBLP:conf/iclr/ShiL023} accelerates model convergence by retrieving discarded messages in backward passes to compute accurate mini-batch gradients. 
\cite{DBLP:conf/cikm/XinSHHQ022} introduces a centrality-based subgraph generation algorithm to solve the problem of information loss among subgraphs caused by the traditional graph sampling process.
\cite{DBLP:conf/asplos/YangZD023} effectively mitigates redundancy and load imbalance between subgraph partitions through both redundant embedded graph partitioning and memory-aware partitioning.
\cite{DBLP:conf/eurosys/HuangZ0WJZZZYS24} introduces a GCN training framework for adaptive joint optimization of graph data partitioning and GCN partitioning to bridge the information gap arising from graph-centric partitioning.

However, the current subgraph sampling methods only have a local receptive field on the subgraph, which makes it difficult to capture the global information in the graph. This limitation significantly impacts the accuracy of GCN in downstream tasks. 
Concerning this, few studies optimize subgraph sampling by integrating external information to reduce the gaps among subgraphs. 
For instance, GAS \cite{DBLP:conf/icml/FeyLWL21} incorporates historical node embedding into the training process of GCN to ensure the integrity of the graph information. LMC \cite{DBLP:conf/iclr/ShiL023} retrieves the neighbor information of the subgraph to expand the receptive field of the subgraph.
These methods represent nodes as free parameters, and different nodes can learn different information during model training. 
As the number of nodes and the complexity of the neural network increase, the model has better expressiveness while incurring high memory consumption. 
\cite{DBLP:conf/icml/FeyLWL21}, \cite{DBLP:conf/iclr/ShiL023}, and \cite{wu2023causality} have shown that gaps in the model can significantly affect the performance of the model.
Therefore, how to bridge the gaps among the subgraphs to improve accuracy while maintaining low memory consumption is an essential problem for GCN.

\subsection{Prompt Learning in Text Pre-train and Fine-tune}\label{R2}
With the rise of pre-trained language models \cite{ding2023parameter} in the field of NLP, fine-tuning methods utilizing prompt learning have demonstrated superior performance on tasks such as intent recognition \cite{DBLP:conf/acl/QiWDC22} and text classification \cite{wang2023text}. 
Prompt learning effectively bridges the gaps between the model and the downstream task. These prompts usually consist of a small number of free parameters and take up only a small amount of memory. Overall, prompt learning is typically divided into two forms: discrete and continuous prompts.

Discrete prompts (i.e., hard prompts) are a method of describing prompts in discrete spaces, typically phrases in natural language that humans can understand. XNLI \cite{DBLP:conf/acl/QiWDC22} introduces a prompting strategy for cross-language natural language understanding tasks, using additional sampling templates to attach a question example to each question. PET \cite{DBLP:conf/eacl/SchickS21} reconstructs the input data as natural language phrases with masked semantics, reconstructing the downstream task of the model. This approach achieves excellent performance on labeled sparse data. 

Continuous prompts (i.e., soft prompts) are a prompting method that performs in the feature embedding space without restricting the prompts to natural language phrases. That is, continuous prompting learns the prompt itself as a task along with the input information. Li et al. \cite{DBLP:conf/acl/LiL20} inserted task-relevant prompt embeddings directly into each layer of the pre-trained model to improve the model's performance on downstream tasks by introducing additional prompt parameters.
 PTCAS \cite{DBLP:journals/isci/ChenSX24} proposes a continuous answer search method based on relation extraction and adds TransH as a knowledge constraint module to enable the model to find the best answer representation through the semantic information of relations. 

However, due to the differences in data distribution between text and graphs, textual prompt templates cannot be directly transferred to graph data. Therefore, some existing studies transfer prompt to graph data (Section \ref{R3}).

\subsection{Prompt Learning in Graph Pre-train and Fine-tune}\label{R3}
Inspired by the application of pre-training models in text area, the use of pre-training models in graph area has become a powerful paradigm. 
For example, \cite{DBLP:conf/iclr/HuLGZLPL20} pre-trained GCN's node-level and graph-level representations to capture both local and global information. In addition, some works \cite{DBLP:conf/cikm/JiangL0S21}, \cite{DBLP:conf/nips/XuCLCZ21} use contrastive learning to enhance the feature representation of downstream nodes.
However, there is a gap between the pre-training content and the downstream target, which will affect their generalizability across tasks.
Recent researches \cite{DBLP:conf/www/MaYLMC24}, \cite{DBLP:conf/kdd/Guo0CLSD23} have shown that prompt learning techniques also exhibit superior performance on cross-task graphs. Specifically, task-specific prompts are designed to prompt downstream tasks to bridge the gaps between pre-training and downstream tasks.

GPPT \cite{DBLP:conf/kdd/SunZHWW22} uses prompts to reformulate the link prediction task into a node classification task, effectively bridging the gaps between the pre-trained model and the downstream task. But GPPT cannot handle different downstream tasks. 
Graphprompt \cite{liu2023graphprompt} unifies pre-training and downstream tasks in a prompt template that relies on learnable prompts to cope with different downstream tasks.
GPF \cite{Fang2023UniversalPT} adopts the same prompt template for all pre-trained models to suggest downstream tasks in an adaptive manner.
Building on this work, 
Sun et al. \cite{DBLP:conf/kdd/SunCLLG23} use prompts for node classification, edge classification, and graph classification, achieving superior performance on multi-task graphs. 

Given that prompts have demonstrated efficacy in bridging the gaps between pre-trained models and downstream tasks, a “prompt” strategy is proposed to bridge the gaps among subgraphs.

\section{Conclusion}
\label{con}
In this paper, we proposed PromptGCN, a novel prompt-based lightweight GCN model, to bridge the gaps among subgraphs. 
First, the learnable prompt embeddings were designed to obtain global information.
Then, the prompts were attached to each subgraph to transfer the global information among subgraphs.
Overall, PromptGCN could be easily combined with any subgraph sampling method to obtain a lightweight GCN model with higher accuracy.
In future studies, we plan to further investigate the design of prompt templates and examine how the quality of these templates impacts model performance. 


%




\ifCLASSOPTIONcaptionsoff
  \newpage
\fi

\bibliography{cas-refs}

\end{document}